\documentclass{article}

\PassOptionsToPackage{numbers, compress}{natbib}

    \usepackage[preprint]{neurips_2023}

\usepackage[utf8]{inputenc} 
\usepackage[T1]{fontenc}    
\usepackage{hyperref}       
\usepackage{url}            
\usepackage{booktabs}       
\usepackage{amsfonts}       
\usepackage{nicefrac}       
\usepackage{microtype}      
\usepackage{xcolor}         
\usepackage{subcaption}
\usepackage[utf8]{inputenc}

\usepackage{booktabs} 

\usepackage{booktabs}
\usepackage{multirow}
\usepackage{caption}
\usepackage{amsmath}
\usepackage{amssymb}
\usepackage{mathtools}
\usepackage{amsthm}
\usepackage{amssymb}
\usepackage{pifont}
\usepackage{footmisc}

\makeatletter
\DeclareRobustCommand\onedot{\futurelet\@let@token\@onedot}
\def\@onedot{.}
\def\eg{\emph{e.g}\onedot}

\makeatother

\usepackage{enumitem}
\usepackage{xspace}
\usepackage{algorithm}
\usepackage{algpseudocode}
\usepackage{hyperref}

\title{A survey on Concept-Based Approaches for Model Improvement}
\author{%
  Avani Gupta, P J Narayanan \\
  CVIT, KCIS\\
  IIIT-Hyderabad, India \\
  \texttt{\{avani.gupta\}@research.iiit.ac.in, pjn@iiit.ac.in} 
}
\begin{document}


\maketitle



\begin{abstract}
     The focus of recent research has shifted from merely improving the metrics based performance of Deep Neural Networks (DNNs) to DNNs which are more interpretable to humans. The field of eXplainable Artificial Intelligence (XAI) has observed various techniques, including saliency-based and concept-based approaches. These approaches explain the model's decisions in simple human understandable terms called \textit{Concepts}. Concepts are known to be the \textit{thinking ground of humans}.
     Explanations in terms of concepts enable detecting spurious correlations, inherent biases, or clever-hans. With the advent of concept-based explanations, a range of concept representation methods and automatic concept discovery algorithms have been introduced. Some recent works also use concepts for model improvement in terms of interpretability and generalization. We provide a systematic review and taxonomy of various concept representations and their discovery algorithms in DNNs, specifically in vision. We also provide details on concept-based model improvement literature marking the first comprehensive survey of these methods.
\end{abstract}

\section{Introduction}
With the increasing use of deep learning in various tasks, there is rapid growth in the XAI literature. One of the most promising methods of XAI is concept-based approaches, which explain model predictions in human-understandable units called \textit{Concepts} \cite{Kazhdan2021IsDA}. Concepts are linked to their origin in NeuroScience Literature, in which they are defined as ideas derived or inferred from facts \cite{donham2010deep} and are known to hold the human's mental world together \cite{murphy2002big}. It has been observed that humans learn by creating concept models of things around them, helping them gather generalized knowledge \cite{lake2017building}. There are various spurious correlations which creep in DNNs due to their black box nature. Model explanations in terms of concepts not only lead to better interpretability but also help to identify these spurious correlations and inherent biases also known as Clever-hans.

In Computer Vision, concept identification started with disentangled representations of latent space \cite{kingma2013auto} where it was observed that similar areas in latent space denote images with some common properties (for example, red cars were found closer). While in Natural Language Processing, Word Embeddings \cite{mikolov2013efficient, pennington2014glove} inspired by the idea that similar words should be grouped together in latent space of model (also known as spatially co-located similarity property) have been widely used. 

The interpretability research aims to explain the model's decisions in simple human understandable terms \cite{zhang2021survey}.
Concepts, by their definition, are humanly understandable and, hence, a good choice for interpretability. Recent XAI literature has observed a surge in concept-oriented explanations. There have been many methods that derive concept-based explanations from pre-trained models (post-hoc), while some do it while training the model (ante-hoc).
\citet{kim2018interpretability} paved the way for post-hoc concept-based explanations of models using Concept Activation Vectors, while a parallel work \cite{fong2018net2vec} explained Convolutional Neural Networks (CNN's) filters in terms of concepts. Many works \cite{bai2022concept,chen2020concept,chang2018concept,bontempelli2022concept} build on the above concept representations in a post-hoc manner. Concept Bottleneck Models (CBMs) \cite{koh2020concept} derive concept explanations ante-hoc, putting the weightage of concept explanations on the model while training. Recent work by \citet{yuksekgonul2022post} extended CBMs to post-hoc explanations. 

Concepts are essential for both human-like learning and better generalization by models. A natural intuition for human-like learning is making the model match the human attention for several tasks as done by \cite{shen2021human, gao2022res, liao2022cnn, he2022efficient, helgstrand2022comparing}. One major limitation of such approaches is the requirement of additional human-level attention data.
 \citet{Chang2018ConceptOrientedDL} formalize \textit{Concept Oriented Deep Learning (CODL)} for human alike learning involving concepts. They argue that CODL overcomes several limitations of deep learning, involving concerns of interpretability, transferability, contextual adaptation, and humongous training data requirements. For the scope of this work, we focus on concept-based literature and refrain from providing generic definitions of interpretability/explainability (or XAI) literature, referring the readers to \cite{zhang2021survey, du2019techniques, linardatos2020explainable, vojivr2020editable, Akhtar2023ASO} which present surveys on XAI methods.
Another concept-inspired learning paradigm is Neuro-Symbolic learning, which involves learning like humans using symbolic reasoning \cite{Mao2019TheNC}.

With Concepts spanning Neuroscience literature to interpretability literature, there are vast representations of concepts varying from concept activation vectors, concept trees, and graphs to concept activation areas in latent space. While some of the methods require concept definitions to be given by the user, several works automate this by extracting concepts from models \cite{ghorbani2019towards, wang2022hint}.

Additionally, some recent methods use concepts for improving model performance by encoding prior knowledge or intervening and correcting the model to ensure it uses correct concepts \cite{Sawada2022ConceptBM,Sawada2022CSENNCS}. We provide a systematic survey knitting this vast literature involving concepts.

Our Key contributions include the following:
\begin{itemize}
\item A systematic survey knitting various concept representations.
\item Taxonomy on concept discovery approaches and their metrics along with commonly used datasets.
\item Finally, a first-of-its-kind survey on CODL techniques. 
\end{itemize}

\section{Related work}
\citet{hitzler2022human} present a survey on concept-based explainability techniques but are restricted to Concept Activation Vector (CAV) \cite{kim2018interpretability} representations. \citet{schwalbe2022concept} survey various visual concept embeddings giving a taxonomy of concept analysis approaches and various datasets used for supervision of concept learning. A parallel work \cite{Holmberg2022MappingKR} also surveys knowledge representations mapping to concepts and classifies concept-based XAI into Associations (what if I see), Interventions (what if I do), and Counterfactuals (what if I had done). 

\citet{gao2022going, weber2022beyond, hase2021can} survey various XAI-based approaches for model improvement.
We categorize XAI-based model improvement in terms of the end goal as \textit{(i)} Enhanced interpretability and \textit{(ii)} Better Generalization. Most of the existing literature focuses on the former (i.e., enhancing interpretability of models) \cite{koh2020concept, kazhdan2020now, alvarez2018towards, teso2019toward, Sawada2022CSENNCS, Sawada2022ConceptBM, stammer2021right, lage2020learning, stammer2022interactive, Teso2022LeveragingEI, Friedrich2022ATT, keswani2022proto2proto, xue2022protopformer, wang2023learning, sacha2023protoseg} while some works address the latter (using interpretability to aid more generalizable models) \cite{Friedrich2022ATT, anders2022finding, bahadori2020debiasing, stammer2021right, song2023img2tab, bontempelli2022concept, shao2022right, shu2019weakly, schramowski2020making, kronenberger2020dependency, Bontempelli2021TowardAU}. Such division concurs with \citet{Holmberg2022MappingKR}, who found that explanations with better understanding significantly differ from explanations with the goal of finding actionable decisions.
\citet{gao2022going} classify XAI-based model improvement in terms of the type of explanation method used while \citet{weber2022beyond} classify based on augmentation made for improvement, which can be over data, loss, gradient, model, etc. Explanation Guided Learning (EGL) involves using explanations for model improvement to meet one or both of the above-listed goals (i, ii). Various local EGL techniques are surveyed by \citet{gao2022going}. Explanatory Interactive Learning (XIL) involves interactively using explanations with human feedback to improve the model. \citet{Friedrich2022ATT} survey various XIL techniques using local XAI methods to remove shortcuts in the models.
eXplanatory Active Learning (XAL) involves using Explanations in an Active Learning framework as surveyed by \citet{ghai2021explainable}.

\citet{hase2021can} survey XAI based model improvement methods in Natural Language Processing  while we focus on such methods in Computer Vision, specifically visual concepts-based model improvement.
\cite{beckh2021explainable} survey model improvement methods aiming better explainability (type ii) using Prior knowledge.
\citet{chang2018concept} provide framework for CODL \cite{Chang2018ConceptOrientedDL} using Concept Graphs.
On the other hand, we provide a novel taxonomy of concept representations and a categorization of various Concept Discovery Algorithms and CODL. We restrict ourselves to visual representations of concepts and show applications of CODL in vision.
We organize our survey based on representations for Concepts used by the existing works followed by concept discovery methods, concept-based model improvement, evaluation \textit{of} discovered concepts, and evaluation \textit{by} concepts. 

\begin{figure}
    \centering
    \includegraphics[width=0.8\textwidth, height=2.5cm]{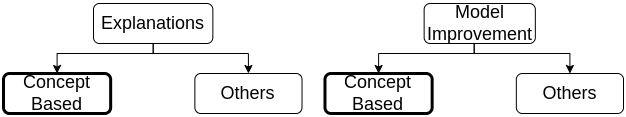}
    \caption{We focus on Concept-based XAI approaches and model improvement approaches}
    \label{fig:teaser}\vspace*{-3mm}
\end{figure}
\section{Various Concept Representations}
The landscape of model interpretability via the lens of concept-based explanations, is rich and varied. This section navigates through various concept representations methods highlighting their unique contributions and applications in deep learning models.

\begin{table}[]
\footnotesize
\tabcolsep=0.04cm
\caption{Concept Representations of existing ML interpretability literature}
\begin{tabular}{|l|l|p{6cm}|}
\hline
\textbf{Concept type}            & \textbf{Concept Representation Method} & \textbf{Papers}  \\ \hline
\multirow{6}{*}{Non-hierarchial} & Hidden Units Concept images alignment based & Network Dissection \cite{Bau2017NetworkDQ}\\
\cline{2-3} 
& Concept Embeddings Vectos based       & CAV \cite{kim2018interpretability}, A-CAV \cite{Soni2020AdversarialT}, CG \cite{bai2022concept}, ICB \cite{zhou2018interpretable}, ICE \cite{zhang2021invertible}, CaCE \citet{Goyal2019ExplainingCW}, ICS \cite{schrouff2021best}, ConceptSHAP 
\cite{Yeh2019OnCC}, \cite{chen2020concept} \\ \cline{2-3} 
                                 & Proto-types & \citet{li2018deep, chen2019looks, bontempelli2022concept}             \\ \cline{2-3} 
                                 & Neuro-Symbolic                         & RRC \cite{stammer2021right}, COOL \cite{marconato2023neuro}, RRR \cite{stammer2021right},
                                 NeuroSC \citet{Marconato2023NeuroSC},
                                 \\ \cline{2-3} 
                                 & Others                                 &   Multi Agent Debate \cite{kori2022visual}               \\ \hline
\multirow{4}{*}{Hierarchial}     & Neuron Attribution                     &        HINT \cite{wang2022hint}         \\ \cline{2-3} 
                                 & Weight Attribution                    & Concept Graphs:  \citet{kori2022interpreting},\citet{Zhang2017KnowledgeVA}, Decision Trees:  POEM\cite{dadvar2022poem}, CHAIN \cite{wang2020chain}   \\ \cline{2-3} 
                                 &  Symbolic features     & Concept trees: \citet{santhirasekaram2022hierarchical},  CNN2DT \cite{Jia2019VisualizingSD}, TreeICE \cite{Mutahar2022ConceptbasedEU},  ACDTE \cite{Shawi2021TowardsAC} \\ \hline
\end{tabular}
\end{table}

\subsection{Non-Hierarchial Concept Representations}
 \subsubsection{\textbf{Hidden Units Concept images alignment based}}
Network Dissection \cite{Bau2017NetworkDQ} uses user-provided concept sets and checks their alignment with individual hidden units at each layer of  CNN. Since DNNs are expected to learn partially non-local representations in denser layers, concepts can align with a combination of several hidden units. However, to assess disentanglement, they focus on measuring the alignment of concepts with single units. This method hypothesizes that the interpretability of units is equivalent to their random linear combination. They evaluate every individual convolutional unit in CNN as a solution to binary segmentation tasks for each visual concept. They pass in all concept sets from the model and determine the distribution of concept activations $a_k$ for each convolutional unit $k$.
They then determine top quantile level $T_k$ for each unit $k$ such that $P(a_k > T_k) = 0.0005$ over every spatial location of the activation map in the concept set. This is followed by a selection of regions exceeding the threshold $T_k$ and evaluating segmentations with every concept $c$ in the concept set by computing the intersection over union ($IoU_{k,c}$) of the above-selected regions with input concept annotation masks. A concept detector is reported if $IoU_{k,c}$ is greater than a certain threshold. The interpretability of a layer is quantified by the number of unique concepts aligned with its units (\textit{unique detectors}). Since IoU is an objective measure (not relative), it enables comparison of interpretability across networks. 
 \subsubsection{\textbf{Concept Embeddings Vectors Based}}
\citet{fong2018net2vec} proposed Net2Vec, which aligns the concepts with CNN filters.
They first collect the pre-trained model's activations for a concept dataset (probe) and then learn the weights to recognize the concept in various semantic tasks. These weights are interpreted as concept embeddings and analyzed to gain insights into how concepts are encoded in the network.

\subsubsection{\textbf{Concept Activation Vectors Based}}
Concept Activation Vectors (CAV) \cite{kim2018interpretability}, a widely popular approach in Concet Based XAI, is encoded as the normal to the boundary separating the activations of the concept of interest and their negative counterpart in the model's activation plane. 
For a human-provided concept of interest, $C$, and its negative counterpart (or random concept) C', a linear classifier is trained to distinguish between the  model activations for a layer $l$ given as $f_l$ for $C$ and $C'$ 
The CAV $v_{C}^l$ is given as a normal to the decision boundary obtained by the above linear classifier.
$v_{C}^l$ gives the direction of the concept in the model's activation plane.
\citet{kim2018interpretability} further uses directional derivatives to gauze the importance given by the model to concept $C$ for the prediction of class $k$.
The \textit{sensitivity} $S_{C,k,l}$ of the model towards $C$ for class $k$ is given as the directional derivative of the model's prediction logit $h_k(x)$ wrt the sample's activation $f_l(\boldsymbol{x})$ in the direction of CAV:
\begin{equation} \label{tcavq}
S_{C,k, l}(\boldsymbol{x}) = \nabla h_{l,k}\left(f_{l}(\boldsymbol{x})\right) \cdot \boldsymbol{v}_{C}^{l}
\end{equation}
The sign of $S_{C, k, l}(\boldsymbol{x})$ tells whether $C$ had an effect over the class sample $x$. This sign is aggregated for all class examples to give the final $TCAV$ score of the model.

Due to their simple and elegant vector-based representation of concepts, CAVs got widely popular and had several works built over them. We now discuss a similar work to CAV followed by their extensions over time.

\paragraph{\textbf{A similar work:  ICB}}Interpretable Concept Basis (ICB) \cite{zhou2018interpretable} uses the weights $w_k$ of the second to last layer (layer before logits) for class of interest $k$, and decomposes them in terms of concept vectors $c_j$ as $w_k \approx \alpha_{k 1} c_1+\alpha_{k 2} c_2+\ldots+\alpha_{k m} c_m$ where $m$ is the total number of user-defined concepts and $\alpha_{k,j}$ are non-negative weights with sparsity constrained to be less than $m$. It thus disentangles and quantifies the contribution of each concept in the model's prediction.

\paragraph{\textbf{ICE: A better approach for fidelity}}
Invertible Concept-Based Explanation (ICE)\cite{zhang2021invertible} proposes Non-negative CAVs (NCAVs) based on Non-negative Matrix Factorization (NMF), which provides better interpretability and fidelity. 
They flatten $f_l(x_c)$ across channel dimention to get $V \in R^{(n \times h \times w) \times c}$.  They use NMF to reduce channel dimensions $c$ to $c'$ and optimize such that $V$ can be approximated as a product of the reduced channel dimension vector $V' \in R^{(n \times h \times w) \times c'}$ and feature direction. 
$$
\min _{S, P}\|V-S P\|_F \quad \text { s.t. } \quad S \geq 0, P \geq 0
$$
Factorization on these vectors helps in disentangling important directions for target concepts. $P$ denotes the meaningful NCAV and is fixed after being trained with $f_l(x_c)$ for some images. Given $P$, NMF can be applied over it to get $S$.
$P$ denotes a vector basis and $S$ is the projection lengths of a vector $A$ onto these basis directions. In this context, $S$ can be interpreted as feature scores that measure the degree of similarity between vector $A$ and the basis directions in $P$. Essentially, $S$ serves as an indication of how much vector $A$ is related or associated with the NCAVs (non-collinear basis vectors) in $P$.
They provide both local and global concept-level explanations for CNN.

\paragraph{\textbf{Extention to Non-linearity}}
CAVs are dependent on the choice of random concept images and the ability of the linear classifiers to distinguish between $C$ and $C'$. Sometimes $C$ and $C'$ can be non-linear, which cannot be handled by CAV. There are two extensions of CAV to non-linearly separable concepts.

\textbf{Adversarial TCAV (A-CAV)} \citet{Soni2020AdversarialT} propose A-CAV, which is more effective in terms of retrieval of concept images using CAVs (improving recall) and is more robust to variations in random examples. 
A-CAV handles the non-linear separability of concepts by pushing the activation vectors away from the decision boundary, adding a small perturbation in the input vector along the direction of its gradient.
A-CAV use two-step Gram-Schmidt Orthogonalization process \cite{bjorck1994numerics} to separate the $f_l(x_C)$ and $f_l(x_C')$ for all $x_C \in C$ and $x_C' \in C'$.
They compute a concept basis, which is the orthogonal basis of subspace-spanned by CAVs, and a non-concept basis, which is disjoint of the concept basis. With concept activations as positive samples, they generate negative samples used for training binary classifiers by sampling activations from a non-concept basis.
To make the CAVs robust to random concept examples, A-CAV trains multiple linear models on different random samples from a non-concept basis and takes the centroid direction of learned coefficients as CAV. Additionally, A-CAV can prevent adversarial attacks and help investigate bias in the model.

\textbf{Concept Gradients (CG)} \citet{bai2022concept} propose CG for surpassing the linearity constraint by CAVs.
Contrary to \cite{kim2018interpretability} which uses model's activation space $R^a$ 
 to calculate CAV's, \citet{bai2022concept} train a concept prediction layer $
g: \mathbb{R}^a \rightarrow \mathbb{R}^c
$ which maps $R^a$ to a different space corresponding concepts $R^c$.
It then calculates the CG as 
$$\mathrm{CG}_{l,k}(f_l(x)) =\nabla g_l(f_l(x))^{\dagger} \nabla h_{l,k}(f_l(x))$$ where, $\nabla g(x)_{l}^{\dagger} \in \mathbb{R}^{m \times d}$ is pseudo-inverse of $g_l(f_l(x))$ and measures the effect of small changes in each concept dimension on model activation $f_l(x)$ while $\nabla h_{l,k}(f_l(x))$ measures the effect small changes in models activations $f_l(x)$ on the final prediction $h_{l,k}$. Thus, the $CG$ measures the concept space changes wrt activation space and output space changes wrt activation space, thereby giving changes in outputs wrt concept (gauging sensitivity of model for concepts).

\paragraph{\textbf{Moving from Correlation measure to Causality}} \citet{Goyal2019ExplainingCW} argue that TCAV scores \cite{kim2018interpretability} can lead to wrong intuitions when concepts are correlated. For example, in the BARS \cite{Goyal2019ExplainingCW} dataset, the vertical and horizontal lines are denoted as two labels, while color is confounded with the orientation of lines. (\eg vertical lines are red, and horizontal lines are green). In this case, even though the model learned to predict the lines, the $TCAV$ score will be high for concept color due to confounding.
\citet{Goyal2019ExplainingCW} overcome this by bringing \textit{Causality} into picture. They generate counterfactual examples of the concept of interest and check the causal effect of concept $c$ given by the Causal Concept Effect (CaCE):
$$\operatorname{CaCE}(\mathbf{c}, f)=\mathbb{E}[f(\mathbf{x} \mid \operatorname{do}(\mathbf{c}=1)]-\mathbb{E}[f(\mathbf{x} \mid \operatorname{do}(\mathbf{c}=0)]$$.

\paragraph{\textbf{Combination with local method }}\citet{schrouff2021best} combine ther global TCAV method \cite{kim2018interpretability} with the local explanations captured by Integrated Gradient \cite{sundararajan2017axiomatic} (IG) to give $TCAV_{ICS}$. 
They take the projection of IG along $v_{c}^l$. Originally, IG used a baseline image $b$ that is conceptually neutral. In object recognition tasks, this image can be white or gray.

$$
\operatorname{ICS}_C^k\left(\mathbf{a}, \mathbf{a}^{\prime}\right):=\left(f_l(\mathbf{x})-\mathbf{a}^{\prime}\right)^T \mathbf{v}_{\mathbf{C}} \int_{\left[\mathbf{a}^{\prime}, f_l(\mathbf{x})\right]} \nabla_{\mathbf{v}_{\mathbf{C}}} h_k(\mathbf{a}) d \mathbf{a}
$$
where $a' = f_l(b)$.
ICS authors also add two new baselines \textit{(i)} Concept forgetting given by $a' = a - \lambda \mathbf{v}_{\mathbf{C}}$ and \textit{(ii)} concept occluding baseline given by $\mathbf{a}^{\prime}=\mathbf{a}-\left(\mathbf{a}^T \mathbf{v}_{\mathrm{C}}+b\right) \frac{\mathbf{v}_{\mathrm{C}}}{\left\|\mathbf{v}_{\mathrm{C}}\right\|_2^2}$. 


\paragraph{\textbf{Completeness via Shapeley values}}
Completeness measures how sufficient a particular set of concepts is for explaining the model's prediction.
ConceptSHAP \cite{Yeh2019OnCC} measures the Completeness of concepts by using Shapeley values associated with cooperative game theory. 
\paragraph{\textbf{Latent Space Concept Disentanglement}}
Concept Whitening \cite{chen2020concept} argues that CAVs might not give a correct picture of concepts' proximity due to latent space being correlated. Thus, they disentangle concepts by decorrelating the latent space and projecting concepts such that they align in different directions, preferably basis directions.


\subsubsection{\textbf{Proto-type based}}
\citet{li2018deep, chen2019looks} proposed Networks that dissect an image by finding prototypical parts and reason over obtained prototypes for classification. They add a new layer after the second last layer to learn proto-types.
\citet{bontempelli2022concept} use interactions with humans for debugging such prototypical networks. 
Humans tell which part-prototypes are to be used or not used, and the model learns based on it. The part-prototypes might not be human interpretable always, and hence \citet{bontempelli2022concept} enable human users to provide additional concepts for supervision. They demonstrate results on both concept-level and instance-level debugging.
The original model sensitivity calculation is done for the final layer (final layer sensitivity to intermediate layer outputs). \cite{gupta2023concept} introduce proto-types for intermediate sensitivity calculations.

\subsubsection{\textbf{Neuro-symbolic}}
\label{neuro_sym}
RRC \cite{stammer2021right}, COOL \cite{marconato2023neuro}, and RRR \cite{stammer2021right} are neuro-symbolic concept learners who use reasoning modules to determine and train for concepts. RRR \cite{stammer2021right} uses a concept embedding module that learns concepts via slot attention \cite{locatello2020object} and a reasoning module that reasons for the concepts. The concept embedding module creates a decomposed representation of input space which can be mapped to concepts, while the reasoning module makes predictions based on the above-obtained concepts.
They stack the reasoning module after the concept embedding module, deriving the reasoning module's explanation from the given final model's prediction output, while the concept module's explanation is derived from the reasoning module's explanation.
This training helps them to come up with decomposed input representations depicting neuro-symbolic concepts learned by the model. For the reasoning module, they use a Set Transformer \cite{lee2019set} to make the reasoning insensitive to the order of concepts and generate the explanations with the help of Integrated Gradients \cite{sundararajan2017axiomatic}. They introduce Neuro Symbolic Continual Learning which involves solving neuro-symbolic tasks mapping sub-symbolic inputs to high-level concepts, and making predictions via reasoning with prior knowledge. They solve catastrophic forgetting with COOL: Concept-level continual Learning, acquiring concepts in a continual manner (retaining over time). Note that concept-level details remain constant over time for various tasks, which is exploited by COOL to avoid catastrophic forgetting.
\subsubsection{\textbf{Multi-agent debate}}
\citet{kori2022visual} pose concept learning as a multi-agent debate problem where two agents come up with arguments (explanations on the selection of specific features in the model's discretized latent space) to support or contradict the classifier's decision.

\subsection{Hierarchical Concept Representations}
\citet{murphy2002big} present a taxonomy of various hierarchies of concepts in the neuroscience literature.
We, on the other hand, review the use of Hierarchical Concept Representations in interpretability literature.

\subsubsection{\textbf{Neuron Attribution based}} 
HINT \cite{wang2022hint} attributes each concept to various neurons, building bidirectional concept-neuron associations. They consider hierarchical relationships between concepts (eg: dog and cat belong to animals) and attribute them to a set of neurons. HINT also indicates how the neurons learn the hierarchical relationships of categories.
This is done by identifying concept responsible regions in the input image followed by constructing a dataset containing a collection of responsible regions $r_e$ for each concept $e \in ep$ and a collection of background regions $r_b*$.
Following this, a concept classifier $L_e$ is trained, which separates concept $e$ from other concepts and a Shapeley value \cite{shapley1953value} based approach to calculate the contributions of each neuron to the concepts. 
$\boldsymbol{r}_{\mathcal{E} \backslash e} \cup \boldsymbol{r}_{b^*}$.
$$L_e(r)=\sigma\left(\boldsymbol{\alpha}^T r\right) \text{  where } r=z_{\mathcal{D}, i, j} \in \mathbb{R}^{|\mathcal{D}|}.$$
$$\phi=\frac{\sum_{\boldsymbol{r}}\left|\sum_{i=1}^M\left(L_e^{\langle\mathcal{S} \cup d\rangle}(r)-L_e^{\langle\mathcal{S}\rangle}(r)\right)\right|}{M\left|\boldsymbol{r}_{\mathcal{E}} \cup \boldsymbol{r}_{b^*}\right|}
$$
Shapeley scores are calculated for all pairs of $e$, and $d$ to obtain a score matrix $\Phi^{nxc}$ where $n$ is the number of neurons and $c$ is the number of concepts in the hierarchy.
Finally, collaborative neurons responsible for a concept $e$ are identified by taking the top $k$ corresponding column highest Shapeley value neurons(top $k$ in $\Phi_{y,e}, y \in \{0,n\}$) 
The above score matrix enables checking for multi-modality of neurons (encodes multiple concepts) or attaining a hierarchy of concepts a neuron is responsible for.

\subsubsection{\textbf{Concept Graphs}}
\label{wt_att_based}
\citet{kori2022interpreting} use concept graphs where the nodes are the concepts and edges depict relationships between them.
They find weights responsible for detecting concepts in the input image by clustering them according to a metric. Then, they use Grad-CAM attributions of clustered weights over input images to determine the concepts detected by them. They also perform significance tests to check the robustness, consistency, and localization of concepts. 
Once concept identification is done, the relationships between them are to be identified to attain a human-understandable trace. This is done using mutual information between pre-interventional and post-interventional feature map distributions. Suppose we want to find if an edge exists between the cluster of a previous layer (say $C2$) and the cluster of the next layer ($B2$). The pre-interventional intervention tells the information flowing from $C2$ to all clusters of the next layer ($B1$, $B2$,..$Bn$).
The post-interventional intervention tells the information flowing from a $C2$ to $B2$. The mutual information between these distributions is used to detect the link $C2->B2$.

 Microsoft Concept Graph (MCR) \cite{ji2019microsoft} contains 5.4 million Natural Language concepts having a hierarchy (sub-concepts)and describes each concept by a set of attributes and an ample relationship space of ("isA," "locatedIn").
MCR is built upon probase \cite{wu2012probase}, which consists of syntactical extractions learned automatically from billions of web pages. 


\citet{Zhang2017KnowledgeVA} learn an explanatory graph from a pre-trained CNN in an unsupervised manner, disentangling different object part patterns in each CNN filter. 
In the explanatory graph, each node represents a part pattern, and each edge encodes co-activation and spatial relationships between patterns. They 
demonstrate that each graph node consistently represents the same object part (concept) through different images.

Pattern-Oriented Explanations of CNN Models (POEM) \cite{dadvar2022poem} identifies patterns of "if Concept $c$ then class $k$ in CNN's. It does concept identification using Network Dissection \cite{Bau2017NetworkDQ} followed by semantic segmentation via Unified Perceptual Parsing \cite{xiao2018unified} involving concept identification with a pre-trained model. POEM attributes concepts to model predictions using three conditions: \textit{(i)} presence of concept in the image, \textit{(ii)} Filter mapping to the concept, \textit{(iii)} significant overlap between spatial concept location in the image and highly activated area in the corresponding filter activation. POEM does Concept Pattern Mining using CART, Explanation Tables, and Interpretable Decision Sets.
Concept-harmonized HierArchical INference (CHAIN) \cite{wang2020chain} evaluates the relationships between CAVs learned in earlier layers vs. later layers, creating a hierarchical inference concept graph from the DNN. 
\subsubsection{\textbf{Symbolic features based: Concept Trees}}
Several approaches to building Concept Tree have been proposed \cite{jiadt, Jia2019VisualizingSD, santhirasekaram2022hierarchical, Mutahar2022ConceptbasedEU, zhang2021invertible}.
\citet{santhirasekaram2022hierarchical} use symbolic features and generate a hierarchy of symbolic rules. Symbols are attained by discretization of the continuous latent space using vector quantization. This is followed by hyperbolic reasoning, which generates an abstraction tree containing symbolic rules and corresponding visual semantics. They demonstrate class-level and instance-level trees.

 CNN2DT \cite{Jia2019VisualizingSD} makes visual surrogate decision trees by decomposing a CNN into a feature extractor and classifier. It identifies concepts by using Network Dissection \cite{}. Then, a decision tree is extracted from the classifier. They provide both local and global interpretations. The DT identified by CNN2DT was found to be too sensitive to training instances.
Additionally, one major limitation of methods using Network Dissection is the requirement for semantic labels for concepts.
 
 A recent work, TreeICE \cite{Mutahar2022ConceptbasedEU} uses NCAVs \cite{zhang2021invertible} to build a decision tree using extracted model features. They first use the ICE algorithm to get concept-level features from CNN. These concept-level features are used to learn NCAVs. Global Average Pooling (GAP) is performed to score each NCAV. A TreeICE classifier is learned using the top-scoring NCAVs and their corresponding concept labels (Ground truth or prediction by CNN).
 
 ACDTE \cite{Shawi2021TowardsAC} extracts concepts and trains a linear model differentiating \textit{clusters of a concept of interest} (unlike concept activations in  CAVs\cite{kim2018interpretability}) and its negative samples (random segments from other clusters). A binary vector $v$ indicating the presence or absence of each concept $c \in C$ is created for each image $s$. Then, a decision tree is learned for image $s$ by using the class predictions and $v$. 
For details on other existing concept representations, please refer \cite{schwalbe2022concept}.

\section{Concept Discovery Methods}
\begin{table}[]
\footnotesize
\tabcolsep=0.04cm
\caption{Concept Discovery categorization}
\begin{tabular}{|l|l|p{8cm}|}
\hline
\textbf{Type}             & \textbf{Method used}                          & \textbf{Papers}   \\ \hline
\multirow{6}{*}{Post-hoc} & Super-pixel                                   & ACE \cite{ghorbani2019towards}, CocoX \cite{akula2020cocox}         \\ \cline{2-3} 
                          & Coorperative Attribution (shapeley)           &  ConceptSHAP \cite{yeh2020completeness}  
  \\ \cline{2-3} 
& Autoencoders based                                   & PACE \cite{Kamakshi2021PACEPA}\\
\cline{2-3} 
& Causality based & MCE \cite{Vielhaben2023MultidimensionalCD}, \citet{Alipour2022ExplainingIC}\\
\cline{2-3} 
& Using Probes & ACDTE \cite{Shawi2021TowardsAC}, \citet{silver2017mastering}\\
\cline{2-3} 
\cline{2-3} 
& Using Convex Optimization & \citet{schut2023bridging}\\
                          \hline
\multirow{5}{*}{Ante-hoc} & Saliency based            &      HINT \cite{wang2022hint}, \citet{kori2022interpreting}           \\ \cline{2-3} 
                          & Using Slot-attention \cite{locatello2020object}                                &  \citet{stammer2021right}                  \\ \cline{2-3} 
                          & Latent Space Disentanglement based & GAN based: EPE \cite{shen2020interfacegan}, StylEx \cite{lang2021explaining}, Dissect \cite{ghandeharioun2021dissect} img2tab\cite{song2023img2tab}, \citet{charachon2022leveraging},  \citet{augustin2022diffusion},  \citet{Tran2021UnsupervisedCB}, Causality enforcing \citet{OShaughnessy2020GenerativeCE}   \\ 
                           \cline{2-3} 
                          & XIL based                                     &  iCSNs \cite{stammer2022interactive}, RRC \cite{stammer2021right}             \\ \hline
\end{tabular}
\end{table}
While \cite{kim2018interpretability} make use of human-supervised concepts, various Concept Discovery Algorithms were developed that automatically discover these concepts.
We classify the Concept Discovery Methods into two categories based on Concept Discovery after model training (Post-hoc) or during it (Ante-hoc).
\subsection{Post-hoc}
\subsubsection{\textbf{Super-pixel based}}
ACE \cite{ghorbani2019towards} and CocoX \cite{akula2020cocox} use super-pixels followed by saliency maps to discover concepts. 
\paragraph{\textbf{Comparing clustered super-pixel segments with a pre-trained classifier}}
ACE focuses on discovering concepts that are meaningful, coherent, and important. On a broad level, ACE uses varying levels of super-pixels to generate both lower (color, texture) and higher-level features (objects) and clusters their activations in the model's activation plane to get concepts. These clusters need to be identified with human interpretable concepts, which require a human to manually go through them. ACE automates this by using ImageNet-trained CNN features as a guide. It compares the perceptual similarity of identified cluster segments with the guide to label them. It also removes outliers to make the concepts \textit{coherent}. It uses TCAV \cite{kim2018interpretability} scores to gauze concept \textit{importance} for prediction. 

\paragraph{\textbf{Using Saliency importances}} For identifying meaningful concepts from super-pixels, CocoX \cite{akula2020cocox} uses Grad-CAM \cite{selvaraju2017grad} followed by pooling technique to obtain importance weights. It then selects the top $p$ super-pixels for each class based on the importance weights and clusters the actual image regions corresponding to important super-pixels for getting concepts. 

\citet{akula2020cocox} further uses counterfactuals for explaining the model's prediction in terms of concepts. It uses fault-line \footnote{Fault lines originate from human cognition where humans zoom in when they imagine an alternative to a
model prediction} identification where adding (or subtracting) a fault line to the concept changes the class of object from A to B. Note that they do not perturb the input image but the activation in the last convolutional layer.

\subsubsection{\textbf{Cooperative attribution based}}
ConceptSHAP \cite{yeh2020completeness} uses Shapeley values that attribute importance to concepts when combined to discover a complete set of concepts.

\subsubsection{\textbf{Autoencoder based}}
PACE \cite{Kamakshi2021PACEPA} extracts concepts via an auto-encoder approach where a 1-D convolutional layer is used to compress different CNN-extracted features. This is followed by an auto-encoder that maps the above-obtained convolutional feature map to concepts and back to the convolutional feature map. They use convolutional and transpose convolutional layers as encoder-decoder to preserve the explanation framework's interpretability.
In order to explain a $K$-way classifier, PACE uses $K$ independent autoencoders and detects the presence of the concept in an image via an embedding map for similarity comparison between the embedding vector and the concept vector.

\subsubsection{\textbf{Causality based}}
Multi-dimensional Concept Discovery (MCD) \cite{Vielhaben2023MultidimensionalCD} proposes concepts that are bound to model reasoning (causality) and are complete. They discover concepts using sparse subspace clustering. \citet{Alipour2022ExplainingIC} present one of few post-hoc methods using generative models for discovering concepts. They use pre-trained generative models to generate global causal contrastive counterfactuals and explain classification models. 

\subsubsection{\textbf{Probe based}}
ACDTE \cite{Shawi2021TowardsAC} finds images similar to class inputs to be explained from a main or related dataset (probe) and segments them. It then clusters the activations of the above-segmented images with few conditions to ensure that segments represent concepts. \citet{mcgrath2022acquisition} delve into how AlphaZero \cite{silver2017mastering}  show that AlphaZero gains human-like chess insights solely via self-play, without learning from human-generated data. The study employs concept-based techniques, showcasing that AlphaZero's neural architecture captures an array of chess knowledge, spanning simple maneuvers to advanced strategies, throughout its training phase. By executing automatic probing, which entails using sparse linear regression models to align the AI's internal activations with established human chess concepts, the study effectively evaluates how these concepts are integrated within AlphaZero's framework. This investigation, complemented by behavioral analysis, pinpoints the timing and location of these concepts within the network, enriching our understanding of AI's potential to mimic human cognitive functions in chess.

\subsubsection{\textbf{Concept vectors search in latent space using convex optimization}}
\citet{schut2023bridging} introduces a method for extracting new chess concepts from AlphaZero, an AI that achieved super-human performance in chess through self-play without human supervision. The study reveals that AlphaZero may harbor knowledge extending beyond existing human understanding, which, however, can be comprehensible and learnable by humans. They use convex optimization to extract learned knowledge explained in terms of concept vectors from a trained AlphaZero model. They optimize $\min \left\|v_{c, l}\right\|_1$ given concept constraints. They use concept constraints for two types of concepts: static and dynamic. Static concepts are found in one state of the RL (\eg, a car located on a highway) model, while dynamic concepts are found in the sequence of states (\eg, accelerating the car). They optimize over the latent vectors to find the best concept vectors for given constraints on concepts. Following concept vector's discovery, they filter concept vectors which are teachable to another AI agent or person. They find AI agents who donot know the concept, teach it to the agent, and evaluate the agent's performance on a task related to the given concept. They use a AlphaZero as a teacher to teach proto-types (chess positions demonstrating the use of concept) to a student agent. For this, they train the student by minimizing the KL divergence between teachers' and students' policies on training proto-types. They then determine whether the concept is teachable by evaluating the student's performance on test set proto-types and estimating how often the student and teacher select the same top-1 move. In order to access the novelty of concepts, they use concepts from the late stages of AlphaZero's training and also introduce a novelty metric based on ease of concept reconstruction using a set of basis vectors from AlphaZero's games. They also provide human evaluation accessing whether the chess grandmasters can learn and apply the discovered concepts. Please refer to the \citet{schut2023bridging} for more details. 

\subsection{Ante-hoc}


\subsubsection{\textbf{Saliency based}}
HINT \cite{wang2022hint} uses saliency maps to identify the responsible regions $r_e$ for concept $e$. 
 They categorize each entry in feature map $z$ as responsible to $e$ or not by aggregating the saliency map $s$ to obtain $\hat{s}$ indicating the relevance of $z_{D}$ to concept $e$. They threshold this saliency map with $t$ to get the responsible regions ($\hat{s} > t$).

GradCAM \cite{selvaraju2017grad} attributions of specific layer's clustered weights are used by Concept Graphs \cite{kori2022interpreting} (discussed in \autoref{wt_att_based}). 

\subsubsection{\textbf{Slot-attention based}}
\citet{stammer2021right} use Slot Attention \cite{locatello2020object} for concept discovery. Slot Attention decomposes the latent space into slots, which are meaningful task-dependent output vectors. Due to this Slot Attention, they can create a differentiable object-centric representation of input image without processing each object of scene \cite{yi2018neural} or dividing the scene into different levels of super-pixels unlike \cite{ghorbani2019towards}.


\subsubsection{\textbf{Using GAN's}}
 EPE \cite{shen2020interfacegan}, StylEx \cite{lang2021explaining}, Dissect \cite{ghandeharioun2021dissect} and img2tab\cite{song2023img2tab} train GANs for learning concepts.
 Dissect \cite{ghandeharioun2021dissect} does concept traversals in increasing order of effect on model prediction. It generates multiple counterfactual examples in input image space.
 
 A recent work img2tab\cite{song2023img2tab} uses StyleGAN features to generate images having certain concepts. It debugs the model for spurious correlations/bias by training it on additional bias-removed data generated by StyleGAN.
\cite{charachon2022leveraging} uses Conditional GANs for generating explanations.

 \citet{Tran2021UnsupervisedCB} discovers causal binary concepts in an unsupervised manner with a VAE. It identifies concepts that are present and concepts that are not present for a model's particular prediction. In their words: "data X is classified as class Y because X has A, B and does not have C" in which A, B, and $C$ are high-level concepts". 
 Their modeling of concepts in this manner encourages causality. \citet{Tonolini2019VariationalSC, Gyawali2019ImprovingDR, Gupta2020PatchVAELL} also use binary concepts but do not impose causality. While \citet{OShaughnessy2020GenerativeCE} uses causal explanations, it does not encourage disentanglement of binary concepts, making the explanations harder to interpret. They use a causal DAG-based representation for modeling concepts. They also propose a learning process to make the model sensitive to the concepts, thus integrating the user's prior knowledge into the model.

 \subsubsection{\textbf{By human interactions}}
 Explanatory Interactive Learning (XIL), as described above, is a learning setting involving interactions with the user for explanations and interventions by them for correcting the model.
 
 \citet{stammer2022interactive} model concept learning as a weak supervision task and use proto-types and XIL for it. They revise the latent space of neural networks with the help of prototypes. For this, they introduce interactive Concept Swapping Networks (iCSNs) which learn concepts and attribute them to specific prototypes by swapping the latent representations of paired images. This results in discrete and semantically representative latent space, which is more interpretable. They also show results on their newly introduced Elementary Concept Reasoning (ECR) dataset, which focuses on visual concepts shared by geometric objects.
 RRC \cite{stammer2021right} uses interactions with humans to improve models' reasoning. These explanations need not be correct because the model can still focus on confounding factors (wrong reasons), and thus, they allow users to correct the explanations by intervening in the model (XIL), giving feedback on explanations of the module which is going wrong. For visual explanations, a user can indicate regions that must be focussed upon \cite{ross2017right, schramowski2020making, teso2019explanatory} in terms of binary masks in input image space.
 For the reasoning module's explanations, user feedback is in the form of relational functions: if concept $C$, then class $k$.
 They regularize the model to match the user explanations via a loss which includes either RRR \cite{ross2017right} or HINT \cite{wang2022hint} loss terms.



\section{Evaluation of Discovered Concepts}
The measures used for the evaluation of discovered concepts are given below:
\begin{itemize}
    \item \textbf{Completeness}:  As introduced by \cite{yeh2020completeness} it measures the sufficiency of a concept in explaining DNN.
    \item \textbf{Correctness/Soundness}: The identified concepts should be correct (not spurious), i.e., they should be truthful to the task model.
    \item \textbf{Fidelity}: It measures completeness and soundness of explanation \cite{kulesza2013too}.
    \item \textbf{Causality}: Concepts \textit{cause} a model to give a certain prediction. This is usually measured by checking the model's behavior on images containing the concept vs images not containing the concept (checked by counterfactual-based approaches \citet{Goyal2019ExplainingCW}.
    \item \textbf{Diversity}: It measures the range of features covered by concepts. Discovered Concepts for a particular input should be non-overlapping.
    \item \textbf{Impurities Measurement} Purity is defined as the degree to which the predictive power of the learned representation for one concept is similar to what we would expect based on its corresponding ground truth label, compared to its predictive power for other concepts  \cite{Zarlenga2023TowardsRM}.
    
    Concept Learning Models are prone to encoding impurities, which are measured by \cite{Zarlenga2023TowardsRM} in the following metrics:
    \begin{itemize}
        \item Oracle Impurity Score (OIS): measures intra-concept impurities while
        \item Niche Impurity Score (NIS) measures inter-concept representations impurities.
    \end{itemize}

\end{itemize}

\section{Different Learning Paradigms for XAI based Model Improvement}
\subsection{Explanation Guided Learning}
Explanation Guided Learning (EGL) is learning that uses Explanations for training or improvising the model. Explanation-guided learning (EGL) can have one of the following aims: (i) improving the interpretability, and (ii)improving the performance of the model. The primary goal of EGL is to learn a model that can make accurate predictions while generating meaningful explanations for its predictions. 
EGL typically involves jointly optimizing the model prediction and the explanation by incorporating three key terms in the objective function: task supervision, explanation supervision, and explanation regularization.
$$\min \underbrace{\mathcal{L}_{\operatorname{Pred}}(f(X), Y)}_{\text {task supervision }}+\underbrace{\alpha \mathcal{L}_{\operatorname{Exp}}(g(f,\langle X, Y\rangle), \hat{M})}_{\text {explanation supervision }}+\underbrace{\beta \Omega(g(f,\langle X, Y\rangle))}_{\text {explanation regularization }}$$
where $\hat{M}$ incorporates the ‘right’ explanation provided via human annotation. The task supervision term guides the model in learning task-specific information, while the explanation supervision term supervises the model explanation to ensure consistency with ground truth. The explanation regularization term helps to avoid overfitting and encourages the model to generate explanations that are interpretable and meaningful.
\begin{figure}[t]
    \centering
\includegraphics[width=0.8\textwidth, height=5cm]{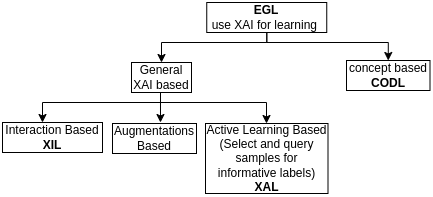}
    \caption{Broad Categorization of Explanation Guided Learning (EGL) methods.}
    \label{fig:EGL_cate}
\end{figure}
EGL is an umbrella that includes XIL, XAL, and CODL approaches \autoref{fig:EGL_cate}.

\subsection{Explanatory Interactive Learning (XIL)}
Explanatory and Interactive Learning (XIL) is a framework for machine learning models to be inspected, interacted with, and revised to ensure that their learned knowledge aligns with human knowledge. XIL methods aim to mitigate learning shortcuts and provide explanations for the model's decision-making process, enabling users to interact with the model and provide feedback to improve its performance. \citet{Friedrich2022ATT} provide a typology of existing XIL methods and provide a generalized XIL algorithm consisting of four essential steps of Select, Explain, Obtain, and Revise. The XIL algorithm takes in a set of annotated examples (A), a set of non-annotated examples (N), and an iteration budget (T). The Select module selects samples from N to present to the teacher. The Explain module provides the teacher with insights into the model's reasoning process. The Obtain module allows the teacher to observe whether the model's prediction is correct or incorrect and to provide corrective feedback. Finally, the Revise module uses the corrective feedback to update the model's behavior towards the user.

\subsection{Explanatory Active Learning (XAL)}
Explanatory Active Learning (XAL) is an emerging learning paradigm that combines active learning with explanations \cite{ghai2021explainable}. \textbf{Active Learning (AL)} is a learning paradigm that allows a learning algorithm to intelligently select instances to be labeled, which can lead to high performance with much less training data compared to traditional supervised learning approaches. AL has become increasingly important in modern machine learning, where labeled data can be expensive and time-consuming to obtain. 
Active Learning reduces labeling workload by selecting instances to query a machine teacher for labels intelligently. However, the human-AI interface remains minimal and opaque, hindering the development of teacher-friendly interfaces for AL algorithms. XAL aims to introduce techniques from the field of explainable AI (XAI) into AL settings to make AI explanations a core element of the human-AI interface for teaching machines. In this paradigm, the teacher should be able to understand the reasoning underlying the model's mistakes during the learning process. Once the model matures, the teacher should be able to recognize its progress to trust and feel confident about their teaching outcome. The teacher here can be a human or a large model.

\subsection{Concept Oriented Deep Learning (CODL)}
 Concept Oriented Deep Learning (CODL) \cite{Chang2018ConceptOrientedDL} involves using concept-level supervision for models to improve model interpretability and performance. The major aspects of CODL, as introduced by \citet{Chang2018ConceptOrientedDL}, include concept graphs, concept representations, concept exemplars, and concept representation learning systems supporting incremental and continual learning. CODL leverages a common or background knowledge base, such as Microsoft Concept Graph, for the framework of conceptual understanding. By focusing on learning and using concept representations and exemplars, CODL is able to address the major limitations of deep learning, including interpretability, transferability, contextual adaptation, and the requirement for a large amount of labeled training data.

\begin{table}[]
\caption{Taxonomy of Concept based model improvement methods}
\footnotesize
\tabcolsep=0.02cm
\begin{tabular}{|p{2.8cm}|l|p{6cm}|}
\hline
\textbf{Model Improvement Goal}          & \textbf{Method}                                     & \textbf{Sub Method: Papers}     \\ \hline
\multirow{5}{*}{Better Interpretability} & \multirow{3}{*}{Concept Conditioned Prediction Based}       & Supervised: CBM \cite{koh2020concept}, CME \cite{kazhdan2020now}    \\ \cline{3-3} 
& & Partially-Supervised: CBM-AUC \cite{Sawada2022ConceptBM}, CBP \cite{grupen2022concept}\\ \cline{3-3} 
                                         &                                           
                                         & Unsupervised: SENN \cite{alvarez2018towards}, XAL based: CALI \cite{teso2019toward}, C-SENN \cite{Sawada2022CSENNCS}, \citet{sarkar2022framework}, intCEM \citet{zarlenga2023learning}, TabCBMs \cite{zarlenga2023tabcbm}    \\ 
                                         \cline{2-3} 
                                         & Concept Reasoning based                             & Neuro-Symbolic: RRC \cite{stammer2021right},  DCR \cite{barbiero2023interpretable} \\ \cline{2-3} 
                                         & \multirow{2}{*}{Interaction based}                  & Human Interaction: \citet{lage2020learning}, NesyXIL \cite{stammer2021right}           \\ \cline{3-3} 
                                         &                                                     & Proto-type based: proto2proto \cite{keswani2022proto2proto}, \citet{xue2022protopformer}, \citet{wang2023learning}, \citet{sacha2023protoseg}\\ \hline
\multirow{8}{*}{Better Generalization}   & \multirow{2}{*}{CAV based}                          & few shot: ClArC \cite{anders2022finding}          \\ \cline{3-3} 
                                         &                                          
                                         & zero shot: Concept Distillation \cite{gupta2023concept}\\ \cline{2-3} 
                                         & Causality Based & \citet{bahadori2020debiasing} \\ \cline{2-3} 
                                         & \multirow{3}{*}{Latent Space Disentanglement based} & Neuro-Symbolic Reasoning Based: NeSyXIL \cite{stammer2021right} \\ \cline{3-3} 
                                         &         & GAN based: Img2Tab \cite{song2023img2tab}                       \\ \cline{3-3} 
                                         &       & XAI based: ProtoPDebug \cite{bontempelli2022concept}, CAIPI \cite{schramowski2020making}, Interactive CBM's \cite{Chauhan2022InteractiveCB}, \citet{Bontempelli2021TowardAU}           \\ \cline{2-3} 
                                         & Probability Distribution based     &          \citet{kronenberger2020dependency}                                   \\ \hline

\end{tabular}
\label{ConceptBased_MI}
\end{table}

\section{Categorisation of Model Improvement Methods}
EGL is an umbrella that includes XAL, XIL, and CODL approaches. We classify the EGL techniques based on their goals in two categories: \textit{(i)} Better Generalization and \textit{(ii)} Better Interpretability. Better Interpretability techniques use explanations for better Interpretability of the model, while better Generation techniques aim to use explanations for the generalizing model.
Note \textit{(i)} includes \textit{(ii)} because using explanations for better Generalization will automatically improve the Interpretability of models.

\subsection{EGL categorization based on goal}
 We categorize XAI-based model improvement in terms of the end goal as aiming \textit{(i)} enhanced interpretability and \textit{(ii)} better generalization. Most of the existing literature focuses on the former \cite{koh2020concept, kazhdan2020now, alvarez2018towards, teso2019toward, Sawada2022CSENNCS, Sawada2022ConceptBM, stammer2021right, lage2020learning, stammer2022interactive, Teso2022LeveragingEI, Friedrich2022ATT, keswani2022proto2proto, xue2022protopformer, wang2023learning, sacha2023protoseg} with some works addressing the latter \cite{Friedrich2022ATT, anders2022finding, bahadori2020debiasing, stammer2021right, song2023img2tab, bontempelli2022concept, shao2022right, shu2019weakly, schramowski2020making, kronenberger2020dependency, Bontempelli2021TowardAU}. Such division concurs with \citet{Holmberg2022MappingKR} that found that explanations with the goals of better understanding have a significant difference from explanations to find actionable decisions.
Note \textit{(i)} includes \textit{(ii)} because using explanations for better Generalization will automatically improve the Interpretability of models.
In this section, we restrict ourselves to Concept-Based Model Improvement Methods.

\subsubsection{Aiming better interpretability}
\label{better_interpretability_goal}
These methods aim for no class accuracy impairment along with enhanced Interpretability of the model and can be categorized broadly based on the type of interpretability methods used as shown in \autoref{ConceptBased_MI}. 

\begin{itemize}
    \item \textbf{Concept Conditioned Prediction Based}
    \begin{itemize}
        \item \textbf{Supervised}
            Concept Bottleneck Models (CBMs) \cite{koh2020concept} use concept mapping as an intermediate step in the model's final prediction. Specifically, they map the input $x$ to a concept $c$ using a concept prediction module $g: x \rightarrow c$, which is then used to predict the target class $y$ using a classification module $f: c \rightarrow y$. This conditioning of the model's prediction over concepts improves the interpretability and generalization of the model. 
            The authors tested CBMs using three different training paradigms: independent training of $g$ and $f$ (with $f$ trained on ground truth concepts), sequential training (where $f$ is given $g$'s predicted concept $\hat{c}$), and joint training (where both $f$ and $g$ are optimized over a joint objective). 
            They found that all three training paradigms produced similar classification accuracies for both concept prediction and target class prediction. However, independently trained models had better test-time interventions.  They intervene not on the actual value of concept $c$ but on the model's concept predictions $\hat c$.
         \item \textbf{Partially-Supervised}
            Concept-based Model Extraction (CME) \cite{kazhdan2020now} approximates DNNs using simpler interpretable models like linear, logistic regression, and decision trees. While CBMs require binary concepts, CME can handle multi-valued concepts and can derive concepts combining multiple layers (unlike TCAV, CBM, \, etc., which require one layer to be chosen). Additionally, CME can train in a partially supervised manner with few labeled and other unlabelled samples. Further, it assumes $k$ different concepts forming concept representation $\mathcal{C} \subset \mathbb{R}^k$ such that every basis vector in $\mathcal{C}$ spans all possible values of a particular concept.
            They define two functions for mapping input to concept space ($p: \mathcal{X} \rightarrow \mathcal{C}$ ) and concepts to predicted class space ($q: \mathcal{C} \rightarrow \mathcal{Y}$ ). CME approximates f with $\hat f$ given as $\hat{f}(\mathbf{x})=\hat{q}(\hat{p}(\mathbf{x}))$
            where $\hat{q}$ and $\hat{p}$ are extracted by CME. 
            For this, they define a function $g^l: \mathcal{H^l} \rightarrow \mathcal{C}$ $\hat g$ and extract it by SSMTL \cite{liu2007semi} using an approximation of k separate tasks (one for each concept). For every concept, $i$ \cite{kazhdan2020now} finds the best layer for predicting the concept by minimizing a loss function $l$
            $l^i=\underset{l \in L}{\arg \min } \ell\left(g_i^l, i\right)$
            and finally approximating $\hat p$ as 
            $$\hat{p}(\mathbf{x})=\left(g_1^{l^1} \circ f^{l^1}(\mathbf{x}), \ldots, g_k^{l^k} \circ f^{l^k}(\mathbf{x})\right).$$
            $\hat q$ is approximated in a supervised manner using a mapped set of concept labels and target labels (class labels)as a training set and fitting using Decision trees or logistic regression. Concept Bottleneck Model with Additional Unsupervised Concepts (CBM-AUC) \cite{Sawada2022ConceptBM} combines CBMs with SENNs to extend CBMs to additional unsupervised concepts along with supervised concepts.
            \citet{grupen2022concept} introduce Concept Bottleneck Policies (CBPs) for enhancing interpretability in multi-agent reinforcement learning (MARL). CBPs use conditioning similar to CBMs but in action space in RL instead of classification space in CBMs. They first predict concepts given a state, then use the predicted concepts to select actions, thus enhancing interpretability.
            \citet{das2023state2explanation} present a novel approach to generating concept-based explanations in sequential decision-making settings, focusing on both improving AI agents' learning rates and enhancing end-user comprehension of AI decisions. They learn a joint embedding model to map state-action pairs to concept-based explanations. These explanations are then used for two primary purposes: informing reward shaping during the agent's training and providing end-users with understandable insights into the agent's decision-making process at deployment.

        \item \textbf{Unsupervised}
            Self-Explaining Neural Networks (SENN) \cite{alvarez2018towards} is a type of Unsupervised method that learns interpretable basis concepts by approximating a model with a linear classifier. They focus on explicitness, faithfulness, and stability via regularizing models. For a linear classifier $f(x)=\theta(x)^T h(x)$ where $x$ is input, $h(x)$ is function mapping input to concepts, and $\theta$ represents model parameters, they enforce the following constraints: \textit{(i)} The output of $f$ is approximately same for two close inputs that are $\nabla_x f(x) \approx \theta\left(x_0\right)$ for all $x$ in a neighborhood of $x_0$; \textit{(ii)} model is linear in terms of concepts; \textit{(iii)} Aggregation function for features shall be generic enough such that it is permutation invariant, isolate multiplicative interactions between concepts and preserve sign and relative magnitude of relevance values $\theta(x)_i$. Finally, they define self-explaining prediction model as: $f(x)=g\left(\theta_1(x) h_1(x), \ldots, \theta_k(x) h_k(x)\right)$ with some additional constraints \cite{alvarez2018towards}. When the above formulation is applied over a neural network, it becomes SENN (the condition being $g$ is continuous over concepts and model weights). SENN can use user-provided concepts and learn new concepts that satisfy \textbf{Fidelity}: concepts persevere relative information and \textbf{Diversity}: concepts for a particular input are non-overlapping.
            For SENN, they learn an autoencoder $h$ and ensure sparsity constraint to increase diversity while using proto-types for interpretations. They use a concept encoder that transforms inputs into concepts, an input-dependent parameterizer for generating relevance scores, and an aggregation function that combines them for the prediction of class labels. The concepts and their relevance predictions form an explanation.
            SENN has reduced interpretability in real-world tasks like autonomous driving, which is overcome by C-SENN \cite{Sawada2022CSENNCS}. C-SENN combines Contrastive learning with concept learning of SENN to improve the discovered concepts and task accuracy.
            
            \citet{sarkar2022framework} use a base encoder (which is the same as the second last layer of the classifier) followed by two branches: one for concept and one for classification. The classification branch resembles the last layer of the classifier and gives the final class label prediction, while the concept head has a concept decoder that reconstructs the image given the intermediate representation (features by the base encoder). Thus, the intermediate representations act as concepts. They additionally impose an image reconstruction loss apart from fidelity and classification losses.

            \citet{zarlenga2023learning} propose intervention Concept Embedding Models (IntCEMs), a novel architecture designed to improve a model's receptiveness to test-time interventions by embedding the capacity for concept interventions directly into the training phase. Unlike traditional Concept Bottleneck Models (CBMs) that lack explicit training for interventions, IntCEMs incorporate an end-to-end learnable intervention policy. This policy predicts meaningful intervention trajectories during training, enabling the model to incorporate expert feedback at test time effectively. 

            Tabular Concept Bottleneck Models (TabCBMs) \cite{zarlenga2023tabcbm}  are designed to generate concept-based explanations for tabular tasks, addressing the gap in concept-based interpretability for non-image data. This model formalizes a high-level concept in tabular data as nonlinear functions of correlated feature subsets, allowing for both supervised and unsupervised concept learning. TabCBMs can learn meaningful, interpretable concepts even without explicit concept annotations, and their performance competes with or outperforms existing methods while providing a high degree of interpretability. The architecture discovers and utilizes concept masks and scores for explanations, supporting human-in-the-loop interventions by allowing experts to correct or modify concept predictions, thereby enhancing model performance. 

        \end{itemize}
    
    \item \textbf{Concept Reasoning Based}
    \textbf{(Neuro-Symbolic)} Right for the Right Concept \cite{stammer2021right} uses a concept embedding module that learns concepts via slot attention Interaction based\cite{locatello2020object} and a reasoning module that reasons for the concepts. The concept embedding module creates a decomposed representation of input space that can be mapped to concepts, while the reasoning module makes predictions based on the concepts that have been identified above. \cite{barbiero2023interpretable} introduces Deep Concept Reasoner (DCR) that uses symbolic reasoning to provide interpretable and semantically consistent predictions. DCR constructs interpretable syntactic rule structures using high-dimensional concept embeddings and then evaluates these rules on semantically meaningful concept truth degrees. This unique approach not only enhances interpretability but also significantly improves model performance on challenging benchmarks, demonstrating the ability to uncover meaningful logic rules even without concept supervision during training. 

    \item \textbf{Interaction based}
    Interaction-based methods aiming for better interpretability need interaction with an expert, which is either done by a human expert or by using class prototypes. Henceforth, they can be classified into broadly two categories as follows.
    \begin{itemize}
        \item \textbf{Human interaction based}
        \citet{lage2020learning} learn interpretable models via user feedback on concepts(which ones are similar and which are not) and also which concepts should(not) affect. 
        Interactive Concept Swapping Networks iCSN's \cite{stammer2022interactive} bind concepts to prototypes by swapping the latent representations of paired images. They use prototypes to interactively learn concepts with user feedback. A Human can query iCSN prototypes and update them for concepts.\\
        NesyXIL \cite{stammer2021right} adds a user interactive layer over Nesy \cite{stammer2021right} discussed above.\\
        For a detailed survey on methods that use explanations in interactive ML, please refer \cite{Teso2022LeveragingEI, Friedrich2022ATT}.

        \item \textbf{Proto-type based}
        Proto2Proto \cite{keswani2022proto2proto} uses knowledge distillation to transfer interpretability from (interpretable) teacher to a student model. 
        
        \citet{xue2022protopformer} use global and local prototypes for enhanced interpretability in Visual Transformers (ViT). \citet{wang2023learning} use macro (broader) and micro proto-types (more specific) for interpretable models learning from mistakes.  A similar idea of macro and micro prototypes is used by \citet{sacha2023protoseg}, which leverages support prototypes capturing macro-level features and trivial proto-types capturing specific micro features. The support (or macro) proto-types provide a global overview of the concepts (or features) but cannot capture some class-specific trivial features  (thus captured by trivial (or micro) proto-types.
    \end{itemize}
\end{itemize}

\subsubsection{Aiming better generalization}
These models typically remove bias or confounding factors and show accuracy improvement on poisoned datasets. \cite{Friedrich2022ATT} gives a topology exploring mitigation of shortcut behavior in models which involves steps of Select, Explain, Obtain feedback, and Revise model.
We divide these methods as follows.
\begin{itemize}
    \item \textbf{CAV based}
    They use CAVs \cite{kim2018interpretability} for the representation of concepts and move activations to make a model sensitive or insensitive to the concept.
    \begin{itemize}
        \item \textbf{Few Shot} \citet{anders2022finding} does artifact removal from models by moving activations of images according to the CAVs learned for class images containing artifact vs. non-artifact class images. It is a few-shot method since it requires artifact and non-artifact images. 
        It uses two methods for the removal of artifacts: Augmentative and Projective. Augmentative Class Artifact Compensation (AClArC) augments the class samples, trying to remove the artifact by moving them according to the CAV direction and retraining the original model. Projective Class Artifact Compensation (PClArC), on the other hand, moves the class activations to a concept-neutral direction in the model's activation space by a simple linear transformation.
        \item \textbf{Zero Shot} \citet{gupta2023concept} introduce a novel method for concept sensitive finetuning of Neural Networks. They introduce a concept loss that moves model activations away from CAV direction for a desensitizing model for a given concept. They further propose \textit{Concept Distillation} to use a pre-trained model's conceptual knowledge to train a student model via their concept loss. They also introduce a proto-type-based concept sensitivity calculation to enable intermediate layer sensitivity. They show applications of their method in debiasing and prior-knowledge reconstruction.
    \end{itemize}
    \item \textbf{Causality based}
Causal modeling involves checking for causes of a particular effect (here, the model's particular predictions). \citet{bahadori2020debiasing} use causal graphs for debiasing CBM's removing confounding factors or clever-hans. It models the impacts of unobserved variables using causal graphs and removes them with a two-stage regression technique aided by instrumental variables.

    \item \textbf{Latent Space Disentanglement Based}
They disentangle latent space to represent similar concepts in similar spatial regions. Latent space disentanglement is typically achieved through the use of generative models, such as autoencoders, variational autoencoders (VAEs), and Generative Adversarial Networks. 
 \begin{itemize}
    \item \textbf{GAN based}
    Img2Tab \cite{song2023img2tab} as described above, allows user interventions on the model for concept based debugging by identifying class-relevant concepts from $W_k$ metric or classifier and using those to filter the semantics our of classifier $P$ by masking all unwanted features across training samples.
    \item \textbf{Neuro-Symbolic Reasoning Based}
    NeSyXIL \cite{stammer2021right} allows for user corrections to its concept embedding module explanations or reasoning module explanations.
    \item \textbf{User Interaction for input based}
    ProtoPDebug \cite{bontempelli2022concept} debugs part-prototype networks using a concept-level debugger with human supervision regarding what part-prototype is forgotten or kept. Right for the Right Latent factors \cite{shao2022right} provides a debiasing approach for generative models via disentanglement of latent space with human feedback. They enforce disentanglement via ELBO loss and a match pairing loss \cite{shu2019weakly}. CAIPI \cite{teso2019explanatory} propose an XIL framework where DNNs query the user while the user explains the queries and corrects the explanation. Right for the Right Scientific Reasons RRSR \citet{schramowski2020making} use CAIPI or RRR \cite{ross2017right} depending on the task and demonstrate results removing clever-hans phenomena. Interactive CBMs \cite{Chauhan2022InteractiveCB} extend CBMs to XIL using an interaction policy that selects which concept labels to be queried from the user to improve prediction maximally. Their policy combines concept prediction uncertainty and the influence of concept on model prediction. \cite{Bontempelli2021TowardAU} introduce a debugging technique for CBM's debugging using human supervision. They can intervene on both concept-level and input-level bugs. CALI \cite{teso2019toward} extends SENN to XAL setting learning SENNs from class labels and explanation guidance by users. 
    \end{itemize}
    \item \textbf{Probability Distribution based}
    \cite{kronenberger2020dependency} uses explanations to verify (accept or reject) the prediction. They show results over the GTSRB dataset \cite{stallkamp2012man}, which consists of 43 different classes showing German traffic light signs. They further use three attributions of varying complexity as concepts:
    \textit{Simple:} Color 
    \textit{Medium:} Primitive Shapes like squares, circles, triangles, octagons, etc.
    \textit{Complex:} Numbers (0-9) and symbols (truck, animal, car, children, bicycle, etc.) 
    The examples above are included as synthetic data.
\end{itemize}

\subsection{EGL in other applications}
In addition to its applications in explainable machine learning, concept-based approaches have also been used in other areas. For example, \citet{Alabdulmohsin2022AdaptingTL} have applied a concept-based approach to domain adaptation, where they aim to adapt a model trained on a source domain to perform well on a target domain with different characteristics. They use concept representations to bridge the domain gap between the source and target domains.
Another example is the work of \citet{McGrath2021AcquisitionOC}, who analyze the knowledge of concepts learned by AlphaZero, an artificial intelligence system that learned to play chess, shogi, and Go at a superhuman level through self-play. They investigate the similarity between the concepts learned by AlphaZero and the concepts understood by human players. They found that the concepts learned by AlphaZero are similar to those learned by humans, but they are more precise and granular.
ProtoMIL is \cite{Rymarczyk2021ProtoMILMI} is used for whole slide image classification and uses Multiple Instance Learning.
XProtoNet \cite{Kim2021XProtoNetDI} uses global and local explanations for Diagnosis in Chest Radiography. \citet{grupen2022concept} proposes the use of concepts for understanding multi-agent behavior. It conditions each agent's actions on concept-based policies proposing Concept Bottleneck Policies (CBPs), achieving interpretability without decreasing performance. They also allow interventions for desired concepts. \citet{schut2023bridging} discover concepts learned in the chess game in  AlphaZero, trying to broaden human knowledge of those concepts. They further showcase that these concepts learned by AlphaZero are learnable and usable by humans. \citet{mood_board_search} uses CAVs for visual searching by finding images within a dataset that align closely with a user-defined mood board\footnote{Mood boards are visual collages that assemble images, materials, text, and other design elements to convey a specific theme, style.}, essentially mapping each mood board with a CAV. 

\section{Concept Based Model Performance Evaluation}
\cite{Khormuji2023APF} provides a Protocol for Evaluating Model Interpretation Methods from Visual Explanations.
\cite{guptaCSMs} provide an evaluation metric called Concept Sensitivity Metric (CSM) for measuring the disentanglement of concepts in multi-branch networks. They use CAV-based sensitivity scores to calculate CSMs. CSMs are essentially a ratio of desired concepts to non-desired concepts. They demonstrate their results on evaluating model disentaglement over an ill-posed under-constrained problem of Intrinsic Image Decomposition (IID). IID involves decomposing an image into two independent variables: Reflectance (R) and Shading (S). By definition, Reflectance is the illumination-invariant component of the scene, while shading is dependent on illumination. They check the sensitivity of the model for concepts of Reflectance properties variance (albedo/material color variance) and Shading properties variance (illumination variance) and provide two CSM scores $CSM_R$ and $CSM_S$ for measuring the quality of R, and S. CSMs are a necessary condition of evaluating IID (since model R-S disentanglement is necessary).
\citet{collins2023human} measure the impact of human uncertainty in the context of concept-based AI models, mainly focusing on models that allow human feedback via concept interventions. Concept-based models, like Concept Bottleneck Models (CBMs) and Concept Embedding Models (CEMs), traditionally assume human interventions are always accurate and confident. However, real-world decision-making involves human error and uncertainty. The study explores how concept-based models respond to uncertain interventions using two novel datasets: UMNIST, which simulates uncertainty based on the MNIST dataset, and CUB-S, a version of the CUB dataset with densely annotated soft labels reflecting human uncertainty.

\section{Future Directions}
There are many future directions for Concept-based Approaches discussed below.
\paragraph{\textbf{Bias Detection}}
Concept Based Approaches are intended to be used for finding confounding factors and clever-hans learned by the models.
While post-hoc explanations require no model architecture changes, they are ineffective for detecting unknown biases in the system \cite{Adebayo2022PostHE}. 
We thus need better automatic bias detection methods using ML interpretability.

\paragraph{\textbf{Robustness}}
\cite{Sinha2022UnderstandingAE} study the robustness of CBMs and SENNs to adversarial perturbations and define many malicious attacks to evaluate this. They also propose training for defense against such attacks. CBMs observe information leakage about data distribution to concept prediction model due to the use of soft concept labels \cite{Mahinpei2021PromisesAP, Margeloiu2021DoCB}. \citet{Lockhart2022TowardsLT} present a framework to mitigate the leaked concept information in CBMs using Monte-Carlo Dropout. GlanceNets \cite{Marconato2022GlanceNetsIL} provides better interpretability by aligning the model's representation and data generation process. It uses disentangled representations and open-set recognition for this alignment and prevents leakage of concepts.

\section{Conclusion}
We discussed various aspects of concept based approaches starting from concept representation methods to various concept discovery methods. We provided a detailed hierarchy of the methods and also discussed the concept evaluation metrics, followed by an explanation of the guided learning hierarchy, focusing on Concept-Oriented Deep Learning (CODL) methods. We classified concept-based model improvement methods on the basis of two aims: better interpretability and better generalization. We also discussed applications of concept-guided methods in fields like domain adaptation, AlphaZero games, and applications in medical domains. In summary, our discussion commenced from post-hoc concept-based interpretability methods, covered ante-hoc concept-based training methods and concluded with concept-based model performance evaluation.

\bibliographystyle{ACM-Reference-Format}
\bibliography{sample-base}


\end{document}